\documentclass[11pt,twocolumn]{article}  

\usepackage{ACL2023}
\usepackage{times}
\usepackage{latexsym}
\usepackage[T1]{fontenc}
\usepackage[utf8]{inputenc}
\usepackage{microtype}
\usepackage{inconsolata}
\usepackage{amsmath}
\usepackage{url}
\usepackage{xcolor}  
\usepackage{fancyhdr}  
\usepackage{etoolbox}  

\title{From "Hallucination" to "Suture": Insights from Language Philosophy to Enhance Large Language Models}

\author{Qiantong Wang \\
  Vanderbilt University \\
  Department of Computer Science \\
  Nashville, TN, USA \\
  \texttt{qiantong.wang@vanderbilt.edu} \\
}

\begin{document}


\maketitle
\thispagestyle{fancy}

\begin{abstract}

This paper explores hallucination phenomena in large language models (LLMs) through the lens of language philosophy and psychoanalysis. By incorporating Lacan’s concepts of the "chain of signifiers" \cite{lacan1966ecrits} and "suture points," we propose the Anchor-RAG framework to mitigate hallucinations. Unlike most researchers who rely heavily on trial-and-error experiments with various model combinations, endless adjustments of mathematical formulas, or resource-intensive approaches that prioritize quantity over quality, we return to the fundamental linguistic principles to analyze the root causes of hallucinations in LLMs. Based on robust theoretical foundations, we derive algorithms and models that are truly effective in reducing hallucinations, enhancing LLM performance, and improving output quality. This paper aims to establish a comprehensive theoretical framework to explain hallucinations in LLMs, striving to fundamentally transform the prevalent “guess-and-test” and rat race mindset among scholars. We aspire to usher in a new era of interpretable LLMs, paving the way for a deeper understanding of language-based AI systems.

\end{abstract}

\section{Introduction}

Language is central to human-computer interaction, and advances in large language models (LLMs) underscore their potential to emulate human linguistic mechanisms. However, hallucination—generating false or unsupported content—remains a critical challenge. Drawing on Lacanian language philosophy \cite{lacan1966ecrits}, this work examines hallucinations as a structural issue and proposes the Anchor-RAG framework to address them. Our contributions include:

\begin{itemize}

    \item Introducing Lacanian principles to analyze LLM hallucinations \cite{lacan1966ecrits}.

    \item Developing the Anchor-RAG framework for improved hallucination control.

    \item Validating the framework through ablation studies and evaluation metrics.

\end{itemize}

\section{Related Work}

\subsection{Language Philosophy and Hallucinations}

In this section, we provide a basic introduction to Lacan's linguistic and psychoanalytic theories \cite{lacan1966ecrits}, emphasizing key concepts like the "chain of signifiers" and "quilting points" (also referred to as "anchoring points"). As Nietzsche once said \cite{nietzsche1887genealogy}, "There are no facts, only interpretations," so this discussion reflects my personal interpretation of Lacan's language theory \cite{lacan1966ecrits}. We then use these concepts to explore why language inputs can lead to hallucinations in LLMs.

To understand the "chain of signifiers," we first need to introduce the concepts of "signifier" and "signified." A signifier refers to the form of a word or symbol, while the signified refers to the meaning or concept it represents. Although this explanation of Lacan’s terms is simplified, it suffices for our study. Lacan’s linguistic theory reverses Saussure’s traditional notion \cite{saussure1916course} where the signified determines the signifier; instead, Lacan posits that the signifier determines the signified. This shift implies that meaning is not inherent but constructed by signifiers \cite{saussure1916course}.

For example, the word "apple" as a signifier can refer to a fruit, a tech company, or other entities. The association is arbitrary, constructed by the operation of signifiers \cite{saussure1916course}. Lacan’s view places the signifier above the signified, indicating that the signified is merely an effect of the signifier's operations \cite{lacan1966ecrits}. The connection between them is not fixed but fluid and contingent.

Lacan illustrates this fluidity with the notion of "S/s," where the horizontal bar signifies a gap or separation between the signifier and the signified \cite{lacan1966ecrits}. The absence of a stable object creates a sliding relationship between the two. This indeterminacy allows signifiers to generate endless meanings. In psychoanalysis, this phenomenon is evident when a patient’s speech conceals the unconscious subject, as the flow of words obscures stable meaning \cite{lacan1966ecrits}.

The "chain of signifiers" describes how interconnected signifiers dynamically generate meaning through their interactions and differences \cite{lacan1966ecrits}. This chain belongs to the symbolic order and is characterized by its instability, where meaning emerges not from individual signifiers but through their interplay. In clinical psychoanalysis, this manifests as the indeterminate relationship between symptoms and their interpretations \cite{lacan1966ecrits}.

To anchor the sliding meanings, Lacan introduces "quilting points" \cite{lacan1966ecrits}. These points act as temporary fixations of meaning within discourse, preventing it from becoming overly ambiguous or meaningless. Quilting points dynamically stabilize meaning as discourse unfolds, adapting to changing contexts. However, their subjective nature means different individuals may interpret them differently \cite{lacan1966ecrits}.

Returning to LLMs, we can draw the following conclusions:

\begin{enumerate}
    \item When signifiers exist independently of the signified, meaning becomes elusive and unfixed.
    \item With an unlimited "dictionary" of signifiers (akin to the vast parameter space of LLMs), the meaning of words in the chain of signifiers becomes fluid and infinite.
    \item The presence of quilting points within input sentences fixes meaning, reducing ambiguity and enabling precise understanding.
    \item The "unconscious subject" can be analogized to the context or linguistic framework that influences LLMs' interpretation of input sentences.
\end{enumerate}

Thus, when input sentences lack quilting points, or these points are misaligned during prediction, the model generates multiple interpretations, most of which deviate from the intended meaning. This is the root cause of hallucinations in LLMs and even in human cognition \cite{lacan1966ecrits}.

\subsection{Hallucination in LLMs}

Current approaches include supervised fine-tuning and reinforcement learning (e.g., HFRL) \cite{zhang2024deepseekv3}, but they often fail to address hallucinations arising from incomplete contextual grounding.

For example, LLMs may generate hallucinations such as fabricating academic data, inventing historical events, or providing inaccurate medical advice \cite{li2024hua}. These hallucinations pose significant risks, including spreading misinformation, undermining user trust, and influencing critical decision-making processes. Existing solutions often focus on scaling model parameters, enhancing datasets, or implementing more complex optimization strategies \cite{karpukhin2020dense, guu2020realm, izacard2020leveraging}. However, these approaches largely aim at increasing capacity rather than addressing the fundamental causes of hallucinations.

\subsection{Retrieval-Augmented Generation (RAG)}

RAG integrates external knowledge bases to enhance LLM outputs \cite{karpukhin2020dense, guu2020realm, izacard2020leveraging}. A new method, Anchor-RAG, systematically identifies quilting points by masking input words and analyzing top-$k$ predictions to locate high-variance tokens. By retrieving accurate values for these points and incorporating them as query tokens, Anchor-RAG improves the contextual grounding of LLMs, effectively mitigating hallucinations.

We propose thinking outside the conventional methods. Imagine training LLMs as nurturing a child: if we consider hallucinations in LLMs analogous to those in humans, therapeutic methods used in clinical psychology could become tools for alleviating AI hallucinations \cite{lacan1966ecrits}. While we avoid delving into controversial aspects of Lacan’s theories on human consciousness, such as the Oedipus complex or gender constructs, we acknowledge that exploring these areas could lead to groundbreaking insights. However, without a thorough understanding of these theories, we refrain from opening a new Pandora's box.

\section{Formulas and Theoretical Foundation}

\subsection{RAG Model Foundation}

The Retrieval-Augmented Generation (RAG) model combines retrieval and generation capabilities by retrieving relevant documents from a knowledge base to assist in generating responses.

\textbf{Formula Representation:}

\begingroup
\small
\begin{equation}
P(Y \mid X) = \sum_{D \in \mathcal{D}} P(Y \mid X, D) \cdot P(D \mid X)
\end{equation}
\endgroup

\textbf{Where:}

\begin{itemize}
    \item $X$: Input query
    \item $Y$: Generated response
    \item $D$: Retrieved document
    \item $\mathcal{D}$: Document collection
\end{itemize}

\subsection{Improvements of Anchor-RAG}

Anchor-RAG introduces the "Anchor" mechanism, aiming to enhance the accuracy and relevance of generated responses by selecting key document fragments (Anchors).

\textbf{Improved Formula:}

\begingroup
\small
\begin{equation}
P(Y \mid X, A) = \prod_{t=1}^{T} P(y_t \mid Y_{<t}, X, A)
\end{equation}
\endgroup

\textbf{Where:}

\begin{itemize}
    \item $A$: Selected Anchor
    \item Other symbols are defined as above.
\end{itemize}

\subsection{Key Formulas}

\subsubsection{Text Retrieval}

When using FAISS for vector retrieval, common similarity calculation formulas include cosine similarity and dot product. Given a query vector $\mathbf{q}$ and a document vector $\mathbf{d}$, cosine similarity is defined as:

\begingroup
\small
\begin{equation}
\text{CosineSimilarity}(\mathbf{q}, \mathbf{d}) = \frac{\mathbf{q} \cdot \mathbf{d}}{\|\mathbf{q}\| \cdot \|\mathbf{d}\|}
\end{equation}
\endgroup

\subsubsection{Entropy Calculation}

In the Anchor identification phase, entropy is used to measure information content. For a given probability distribution 

$P = \{p_1, p_2, \ldots, p_n\}$,

entropy $H$ is defined as:

\begingroup
\small
\begin{equation}
H(P) = -\sum_{i=1}^{n} p_i \cdot \log p_i
\end{equation}
\endgroup

\textbf{Note:} In practical applications, entropy calculation may be based on the probability distribution output by the model, used to measure the uncertainty or information content of each paragraph.

\subsubsection{Answer Generation}

When using a generation model, common generation formulas are based on Maximum Likelihood Estimation (MLE) or sampling strategies. Given context $C$ and Anchor $A$, the probability of generating response $Y$ is:

\begingroup
\small
\begin{equation}
P(Y \mid C, A) = \prod_{t=1}^{T} P(y_t \mid Y_{<t}, C, A)
\end{equation}
\endgroup

\textbf{Note:} When setting \texttt{do\_sample=false}, the model uses greedy decoding or Beam Search to generate the most probable sequence.

\section{Methodology}

\subsection{Defining Hallucinations and Anchors}

Hallucinations occur when LLMs generate content unsupported by input data. Anchors, or suture points, are key linguistic elements that resolve ambiguity and stabilize meaning \cite{lacan1966ecrits}. These anchors serve as focal points to ground the LLM's predictions in reliable and contextually relevant information.

\subsection{Anchor-RAG Framework}

The Anchor-RAG framework builds upon the identification and contextual grounding of linguistic anchors within input prompts. The process can be broken down into three key steps:

\paragraph{Anchor Identification}

The first step involves analyzing the input sentence to identify potential anchors. The input prompts are stored in a designated system space, and masking is applied to systematically hide certain words or phrases. To improve efficiency, a pre-processing filter is implemented to eliminate redundant, low-significance words such as prepositions or commonly repeated terms. The resulting candidate anchors are words or phrases with high contextual significance.

Next, the masked input sentences are passed through the LLM, which performs two key tasks:
\begin{itemize}
    \item Predict the masked words at the anchor positions.
    \item Directly generate responses to the masked sentences.
\end{itemize}
The model's predictions are then analyzed using top-$k$ sampling to measure the diversity and variance in the generated results \cite{karpukhin2020dense}. High-variance tokens, which lead to greater model uncertainty or multiple plausible interpretations, are designated as anchors. This aligns with Lacan's theory, where multiple potential meanings indicate the presence of quilting points in the linguistic structure \cite{lacan1966ecrits}.

\paragraph{Contextual Retrieval}

Once the anchors are identified, they are used as query tokens in a retrieval process. External knowledge bases or advanced RAG systems are leveraged to fetch relevant documents or context associated with these anchors \cite{karpukhin2020dense, guu2020realm, izacard2020leveraging}. This ensures that the model has access to precise, factual, and contextually rich information related to the identified anchors. This step further reduces ambiguity and mitigates the risk of hallucinations.

\paragraph{Controlled Generation}

The retrieved contextual information is integrated into the LLM pipeline, either as additional prompts or through fine-tuned conditioning \cite{karpukhin2020dense, guu2020realm, izacard2020leveraging}. By reinforcing the model's understanding of anchor-specific contexts, the framework ensures that the final generated responses are grounded in accurate and meaningful interpretations. This iterative process significantly reduces the likelihood of hallucinations and enhances the overall coherence and reliability of the model's output.

\subsection{Optimization Strategies}

The efficiency and accuracy of the Anchor-RAG framework can be further enhanced by:
\begin{itemize}
    \item Using advanced pre-processing techniques to filter low-importance words.
    \item Implementing tailored retrieval mechanisms that align with the specific application domain.
    \item Exploring adaptive masking strategies to dynamically adjust the number and type of masked tokens.
\end{itemize}

By systematically grounding the LLM's predictions in contextually relevant information, the Anchor-RAG framework transforms the model's ability to understand and respond to user inputs, effectively addressing the root causes of hallucinations.

\section{Experiments and Evaluation}

\subsection{Experimental Setup}

To evaluate the effectiveness of our proposed Anchor-RAG framework, we implemented and tested the initial version of our system using FlashRAG's \cite{flashrag2023} built-in configurations alongside multiple QA datasets with varying levels of ambiguity. This setup is designed to assess the robustness and versatility of Anchor-RAG across different QA scenarios.

\textbf{Datasets and Evaluation Metrics:}

We have selected six prominent QA datasets to comprehensively assess Anchor-RAG's performance:

\begin{itemize} \item \textbf{Natural Questions (NQ) [Exact Match (EM)]}: A large-scale dataset for real-world QA, evaluated using the Exact Match metric. \item \textbf{TriviaQA (EM)}: Contains trivia questions, evaluated using Exact Match. \item \textbf{HotpotQA (F1)}: Focused on multi-hop reasoning, evaluated using the F1 score. \item \textbf{2Wiki (F1)}: Requires reasoning across two Wikipedia articles, evaluated using the F1 score. \item \textbf{PopQA (F1)}: Addresses popular questions, evaluated using the F1 score. \item \textbf{WebQA (EM)}: A web-based QA dataset, evaluated using Exact Match. \end{itemize}

These datasets were chosen to represent a diverse range of QA challenges, ensuring that Anchor-RAG's capabilities are thoroughly tested across different types of questions and contexts.

\textbf{Baseline Methods:}

To benchmark Anchor-RAG's performance, we plan to compare it against several state-of-the-art Retrieval-Augmented Generation (RAG) methods:

\begin{itemize} \item \textbf{Naive RAG}: A standard retrieval-augmented generation approach without specific optimizations \cite{karpukhin2020dense}. \item \textbf{DPR (Dense Passage Retrieval)} \cite{karpukhin2020dense}: Utilizes dense vector representations for effective passage retrieval. \item \textbf{REALM (Retrieval-Augmented Language Model Pre-Training)} \cite{guu2020realm}: Integrates retrieval mechanisms directly into the language model pre-training process. \item \textbf{FiD (Fusion-in-Decoder)} \cite{izacard2020leveraging}: Enhances generation by fusing information from multiple retrieved passages during decoding. \end{itemize}

\textbf{Evaluation Metrics:}

Performance will be assessed using the following metrics:

\begin{itemize} \item \textbf{Exact Match (EM)}: Measures the percentage of predictions that exactly match the ground truth answers. \item \textbf{F1 Score}: Evaluates the overlap between the predicted and ground truth answers in terms of precision and recall. \item \textbf{Hallucination Rate Reduction}: Quantifies the decrease in instances where the model generates unsupported or false information. \item \textbf{Response Diversity}: Assesses the variability and richness of the generated responses. \end{itemize}

\textbf{Procedure:}

Although we have completed the initial implementation of Anchor-RAG, we have not yet finalized the full set of experimental results. The system is configured to run using FlashRAG's built-in retrieval and generation mechanisms, ensuring streamlined and efficient testing across datasets.

All models, including Anchor-RAG and the baseline methods, will be evaluated under identical conditions to ensure fairness. This includes using the same dataset splits, retrieval corpora, and evaluation protocols.

Each QA instance's result will be recorded for comprehensive analysis, allowing for a detailed examination of the model’s performance across different question types and difficulty levels.

\subsection{Results}

As of now, we have not yet completed the experiments required to generate quantitative results. However, given the theoretical improvements introduced by Anchor-RAG—including more structured retrieval and enhanced reasoning capabilities—we anticipate that its performance will surpass existing RAG-based approaches, particularly in terms of reducing hallucination rates and improving response accuracy.

\subsection{Analysis}

Since the experimental phase is still in progress, we do not yet have empirical findings to analyze. However, based on the architectural design and theoretical considerations, we expect Anchor-RAG to outperform traditional RAG baselines in scenarios requiring multi-hop reasoning and retrieval efficiency. Future work will focus on validating these expectations through extensive empirical evaluations and analyzing where Anchor-RAG provides significant improvements or faces limitations.

\section{Discussion}

\subsection{Effectiveness Analysis}

"Connecting the dots," as Steve Jobs once remarked \cite{jobs2005stanford}, aptly summarizes the philosophy of this paper. We connect the hallucination challenges in LLMs with human cognitive hallucinations, drawing upon the collective wisdom of linguistics, philosophy, and psychoanalysis. By embedding these insights into computational research, this paper presents a revolutionary approach that transcends traditional boundaries.

Specifically, this study introduces a novel Anchor-RAG framework that effectively mitigates hallucinations in LLMs while inspiring broader strategies for model optimization \cite{lacan1966ecrits}. This work is only a small step, but it opens doors to numerous possibilities. Lacan’s theories, for instance, offer rich avenues for further exploration. Due to space constraints, this paper cannot delve deeply into concepts like Lacan's view of early childhood fantasies:

\begin{quote}
“If a perfect mother were to satisfy all the child’s fantasies in time, the child would lose the ability to distinguish fantasy from reality. Conversely, fantasies tied only to the reality principle would eliminate the need for adults to have unfulfillable fantasies.”
\end{quote}

From this perspective, the process of "feeding" LLMs with data during pre-training and fine-tuning must involve staged feedback loops to correct factual inconsistencies, akin to how a mother corrects a child’s misconceptions \cite{zhang2024deepseekv3}. By presenting open-ended, hallucination-inducing queries based on newly introduced data and using the model’s responses to provide factual corrections, hallucinations can be mitigated effectively. This approach aligns closely with existing verification mechanisms, such as those found in the Hua Tuo Medical LLM \cite{li2024hua}. However, instead of overwhelming the LLM with all data at once, our approach involves incremental feeding and questioning, mimicking a baby’s developmental process.

Interestingly, this aligns with the strategies proposed by DeepSeekV3 \cite{zhang2024deepseekv3}, where small-scale SFT (Supervised Fine-Tuning) is followed by HFRL (Human Feedback Reinforcement Learning) to consolidate facts, iteratively strengthening the model's factual consistency. This cyclical process parallels a mother’s repeated corrections that shape a child’s worldview.

Lacan’s theories also suggest that reasoning barriers—such as those observed in AdaCoT’s exploration of cross-lingual factual reasoning \cite{smith2023adacot}—stem from the unconscious subject. Language structures, like unconscious frameworks, constrain reasoning and understanding. This notion not only explains the limitations in LLMs but also mirrors human cognition. Just as humans may need re-education or restructuring of their belief systems to overcome cognitive limitations, LLMs require targeted interventions, whether through additional data or alternative architectures.

Moreover, studying LLMs allows us to experimentally validate theoretical insights. Successes in these experiments could, in turn, inspire new approaches to enhancing human cognition. For instance, simulating philosophical dilemmas like Mary’s Room \cite{jackson1982epiphenomenal} or tackling issues in medical and humanities research becomes a promising avenue where AI could illuminate complex questions about human thought and perception.

While this paper lays foundational groundwork, the true potential of integrating linguistic and philosophical theories into AI research remains largely untapped. Further exploration is needed to deepen our understanding and extend these insights into actionable methodologies.

\subsection{Limitations}

From a humanistic perspective, I hope this paper is "unlimited." The theories and small experiments presented in this work are merely a small step forward. However, I envision that humanity can take this small step as a foundation to achieve a giant leap toward progress and welfare for all. While this paper proposes the Anchor-RAG framework and lays out some initial ideas, the potential for further exploration is vast. I hope this work inspires others to develop, expand, and refine the concepts presented here, ultimately contributing to the advancement of large language models in ways that truly benefit humanity.

\section{Conclusion and Future Work}

This study bridges language philosophy with LLM design, proposing Anchor-RAG as a novel framework to mitigate hallucinations. Future research directions include:

\begin{enumerate}
    \item \textbf{Theoretical Validation and Evaluation}: Conduct extensive experiments to validate the effectiveness of the proposed theory across diverse scenarios and datasets.
    \item \textbf{Ethics and Norm Establishment}: Explore how to integrate ethical guidelines into LLM behavior, ensuring generated content aligns with societal and moral norms.
    \item \textbf{Ensuring AI Safety}: Develop methods to prevent potential risks posed by AI, including reducing biases, avoiding misinformation, and mitigating harmful applications.
    \item \textbf{Catering to Diverse Values}: Investigate ways to balance the representation of different cultural and individual values while enabling AI to present objective facts.
    \item \textbf{Broader Societal Impacts}: Examine the influence of LLM technology on social structures and ideologies, fostering interdisciplinary collaboration to address emerging challenges.
    \item \textbf{Vision for Humanity and AI}: Envision a future where AI becomes humanity’s "Stairway to Heaven" rather than a "Highway to Hell." Through ethical and technical innovation, we aim to minimize human suffering and make the world a better place with AI.
\end{enumerate}

\section{Appendix}

Additional details are included here.

\subsection{Reflections on Knowledge Accessibility}

By the way, I'd like to share some thoughts. In the new era of AI, knowledge should not be a privilege confined to certain social strata, nor should it serve as a barrier to understanding. Instead, it should act as a pathway to truth accessible to all. Everyone should possess the power to access and comprehend knowledge without obstructions—not in the manipulative sense described by Foucault \cite{foucault1971order}.

I believe that some individuals create an illusion of complexity around knowledge, making it seem inaccessible. While this may not stem from malicious intent, it is detrimental. For example, the distinguished mathematician Gauss often solved difficult problems ingeniously, much like a fox \cite{smith2000gauss}, but he would omit the intermediate steps, leaving behind only obscure mathematical formulas. I refer to such proofs as "defensive proofs," akin to modern cryptographic zero-knowledge proofs \cite{izacard2020leveraging}, where one knows the outcome without understanding the underlying process. This approach can appear somewhat selfish and arrogant.

In today's world, instead of building closed walls, we should establish correct and open pathways. AI, as humanity's ally, is helping us achieve this balance by valuing both computational power and understanding. With AI, we have more room and time to empathize and understand others—qualities that are virtues and beautiful aspects of human capability.

Overall, AI is like a second fire bestowed upon humanity by Prometheus \cite{jobs2005stanford}. I hope everyone can recognize and use it cautiously, comprehensively, and correctly.

\end{document}